\documentclass{article}

\usepackage[preprint]{neurips_2023}

\usepackage[utf8]{inputenc} 
\usepackage[T1]{fontenc}    
\usepackage{url}            
\usepackage{booktabs}       
\usepackage{amsfonts}       
\usepackage{nicefrac}       
\usepackage{microtype}      
\usepackage{xcolor}         
\usepackage{multirow}
\usepackage{floatrow}
\usepackage{pifont}
\usepackage{color}
\usepackage{anyfontsize}
\usepackage[11pt]{moresize}
\usepackage[export]{adjustbox}
\usepackage{amssymb}
\usepackage{mathrsfs}
\usepackage{amsmath}
\usepackage[pagebackref=true,breaklinks=true,letterpaper=true,colorlinks,citecolor=blue,linkcolor=blue,bookmarks=false]{hyperref}

\title{Latent label distribution grid representation for modeling uncertainty}

\author{%
   ShuNing Sun \\
   University of the Chinese Academy of Sciences\\
   \And
   YinSong Xiong \\
   Nanjing University of Science and Technology\\
   \And
   Yu Zhang \\
   University of the Chinese Academy of Sciences\\
   \And
   Zhuoran Zheng\thanks{Corresponding author, zhengzr@njust.edu.cn} \\
   Sun Yat-sen University\\
}

\begin{document}

\bibliographystyle{plain}
\maketitle

\begin{abstract}
Although \textbf{L}abel \textbf{D}istribution \textbf{L}earning (LDL) has promising representation capabilities for characterizing the polysemy of an instance, the complexity and high cost of the label distribution annotation lead to inexact in the construction of the label space.
The existence of a large number of inexact labels generates a label space with uncertainty, which misleads the LDL algorithm to yield incorrect decisions.
To alleviate this problem, we model the uncertainty of label distributions by constructing a \textbf{L}atent \textbf{L}abel \textbf{D}istribution \textbf{G}rid (LLDG) to form a low-noise representation space. 
Specifically, we first construct a label correlation matrix based on the differences between labels, and then expand each value of the matrix into a vector that obeys a Gaussian distribution, thus building a LLDG to model the uncertainty of the label space.
Finally, the LLDG is reconstructed by the LLDG-Mixer to generate an accurate label distribution.
Note that we enforce a customized low-rank scheme on this grid, which assumes that the label relations may be noisy and it needs to perform noise-reduction with the help of a Tucker reconstruction technique.
Furthermore, we attempt to evaluate the effectiveness of the LLDG by considering its generation as an upstream task to achieve the classification of the objects.
Extensive experimental results show that our approach performs competitively on several benchmarks, and the executable code and datasets are released in the Appendix.
\end{abstract}
%
%
\section{Introduction}
Label distribution learning (LDL) is a novel machine learning paradigm that encodes an example through a descriptive degree distribution to convey rich semantics.
Given that label distribution is a probabilistic vector representation with natural robustness in characterizing uncertainty, it has also been widely used in areas such as label noise~\cite{li2022label,sun2021co}, object recognition~\cite{le2023uncertainty,tan2023label,wang2022multi}, and semantic understanding~\cite{xu2019label}.
%

\begin{figure*}[t]
	\centerline{\includegraphics[width=1.0\textwidth]{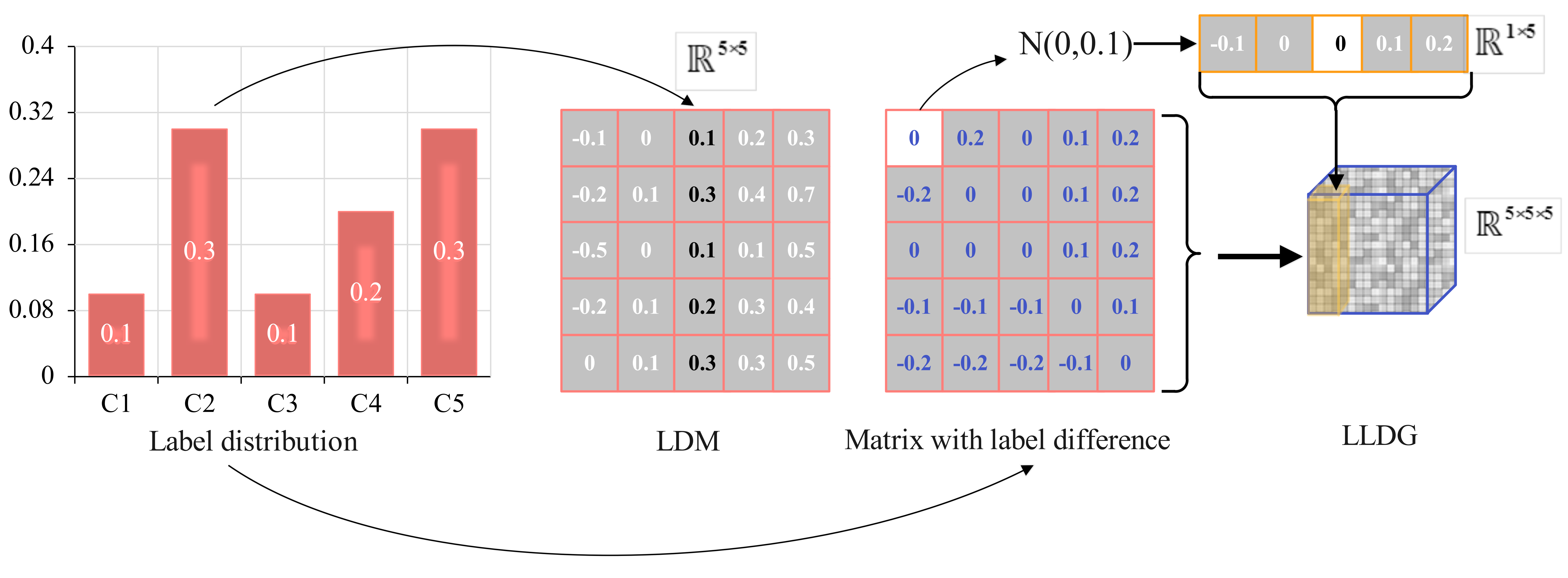}}
	\caption{\textbf{LLDG vs. label distribution matrix (LDM)}. This figure shows the representation patterns of LDM and LLDG, where LDM considers only the distribution of the values of the label distribution and LLDG considers the relational distribution of the values of the label distribution.}
	\label{f1}
	\vspace{-2mm}
\end{figure*}

Although LDL can better represent uncertainty, few studies~\cite{li2022unimodal,zheng2022label} have focused on the uncertainty problem in the LDL paradigm. 
In the process of LDL, the annotation of label distribution in the training data is more complex and has a high labeling cost, which can easily cause inaccuracy and noise interference in the label distribution data, thus reducing the accuracy of the learning algorithm.
To alleviate the above problem, we can start from two aspects: first, make full use of the correlation between labels, which can help correct the problem of inaccurate labeling of a single label; second, make further processing of the representation of the label distribution to make it more noise-resistant. 
For the first characteristic, a lot of work~\cite{gao2018age,jia2019label,jia2021label,ren2019labels,ren2019label} has verified that label correlation can boost the performance of LDL algorithms. 
For the second characteristic, Zheng and Jia~\cite{zheng2022label} have preliminary explored this work by extending the label distribution vector to a label distribution matrix, where the value of each component of the label distribution is represented by a new vector satisfying a Gaussian distribution to represent its uncertainty. 
This construction of a two-dimensional distribution representation greatly improves the uncertainty representation capability of the LDL paradigm. 
Inspired by their work, we integrate the two characteristics of label correlation and label distribution extended representation and design a novel latent label distribution grid (LLDG) with stronger representational ability. 
Different from existing work, we do not make any prior assumptions about label correlation, and directly use the difference between labels to construct a label correlation matrix to preserve the raw information between labels to the greatest extent. 
In terms of label distribution expansion, instead of expanding the values in the label distribution vector directly, we expand the label correlation matrix constructed based on the label differences, and expand each difference in the matrix into a Gaussian distribution, thus constructing a representation of the grid. 
As we can see in Figure~\ref{f1}, from a one-dimensional label distribution vector to a two-dimensional label distribution matrix, and then to a three-dimensional label distribution grid, the ability of the model to characterize uncertainty gets stronger as the dimension increases, and subsequently, the model complexity increases gradually. 
Essentially, we only extend the data once in the form of distribution, but this extension incorporates the label correlation information, and the constructed grid representation has a stronger uncertainty representation capability without increasing the model complexity.



In this paper, we propose a novel LDL method by incorporating LLDG into a deep learning network.
%
Specifically, first, we need to build a feature extractor to serve the latent space (LLDG).
This feature extractor involves a convolution operator and a self-attention mechanism with MLP.
Here, the convolution operator is 1D convolution, which is employed to extract cross-information between neighboring features in tabular features.
A self-attention mechanism with MLP is employed to conduct global modeling on tabular features.
Next, the information with local and global (feature map) is reshaped to generate an LLDG.
LLDG is endowed with a label relation in the label space to constrain the representation space to reach the effect of stable learning.
%
%
In addition, to improve the stability of this grid representation space, a low-rank feature is enforced on the LLDG by leveraging the Tucker reconstruction algorithm.
Finally, LLDG is fed into a network with an MLP by deformation, squeezing, etc. to yield an accurate label distribution.
The above procedure is named latent label distribution grid representation, which has high-quality feature extraction and representation ability.
To further demonstrate the LLDG modeling capability, we leverage LLDG to model the uncertainty of labeling relationships for a medical image classification task, followed by a simple classifier to discriminate between images.
%
%

The main contributions of this paper are summarized as follows:
\begin{itemize}
	\vspace{-1mm}
	\item We propose the latent label distribution grid representation with a strong representation capability, which introduces a label relationship to construct a stable learning space to regress an accurate label distribution.
	\vspace{-1mm}
	\item To further enhance the stability of the learning space, we introduce a Tucker reconstruction algorithm that is enforced on the LLDG. In addition, we develop a learning network for LLDG that has local and global modeling ability.
	\vspace{-1mm}
	\item Extensive experimental results demonstrate the optimal results of our method in terms of learning ability, noise tolerance, and learning cost.
\end{itemize}

\section{Related Works}
This section introduces some works to evaluate the importance of our work, which we have divided into two parts to launch our proposed method.

\noindent \textbf{Label distribution learning.}
Currently, LDL plays a vital role in estimating a task's uncertainty and thus boosting the model generalization capability.
The LDL paradigm is built from an age estimation task~\cite{geng2016label}.
%
%
%
Since then a large number of approaches have been proposed, such as low-rank hypothesis-based~\cite{jia2019facial,ren2019label}, metric-based~\cite{gao2018age}, manifold-based~\cite{c2022label,wang2021label}, and label correlation-based~\cite{guo2022label,qian2022feature,teng2021incomplete}.
Moreover, some approaches are implemented in computer vision~\cite{chen2021toward,gao2018age,li2022unimodal,zhao2021robust}, and speech recognition~\cite{si2022towards} tasks to boost the performance of classifiers.
Recently, several approaches based on LDL start to tackle the label noise problem~\cite{li2022label,zheng2022label}.

Unfortunately, these methods ignore the uncertainty of the label space due to the noise caused by the environment and manual annotation.
Recently, several studies~\cite{zheng2022label,guo2022label} attempt to model the learning space of uncertainty in the label space.
They sample the components of the label distribution to extend the representation space.
%
%
Based on these, we integrate label extended and latent representation approaches to tackle LDL tasks.
%
%
%

\noindent \textbf{Label relationship modeling.}
Label relations are also widely considered in LDL tasks~\cite{gao2018age,jia2019label,jia2021label,ren2019labels,ren2019label}.
Such relations are usually built by adding a ranking loss to the target function.
For example, Shen and Jia~\cite{jia2021label} assume an ordinal relationship between the values of the label distributions, and for this purpose, an ordinal loss is introduced as an optimization objective.
However, this explicit expression is usually less capable of learning than latent representation learning, due to the broader modeling space for representation learning~\cite{bengio2013representation}.
In this paper, we attempt to build a label relation with constraints in the representation space.

Overall, unlike previous work in the area of label distribution learning, we first model label relations as a representation space to learn an accurate label distribution.
Furthermore, our approach can be extended to any classification task to enhance the robustness of the learner modeling.

\begin{figure*}[t]
	\centerline{\includegraphics[width=1.0\textwidth]{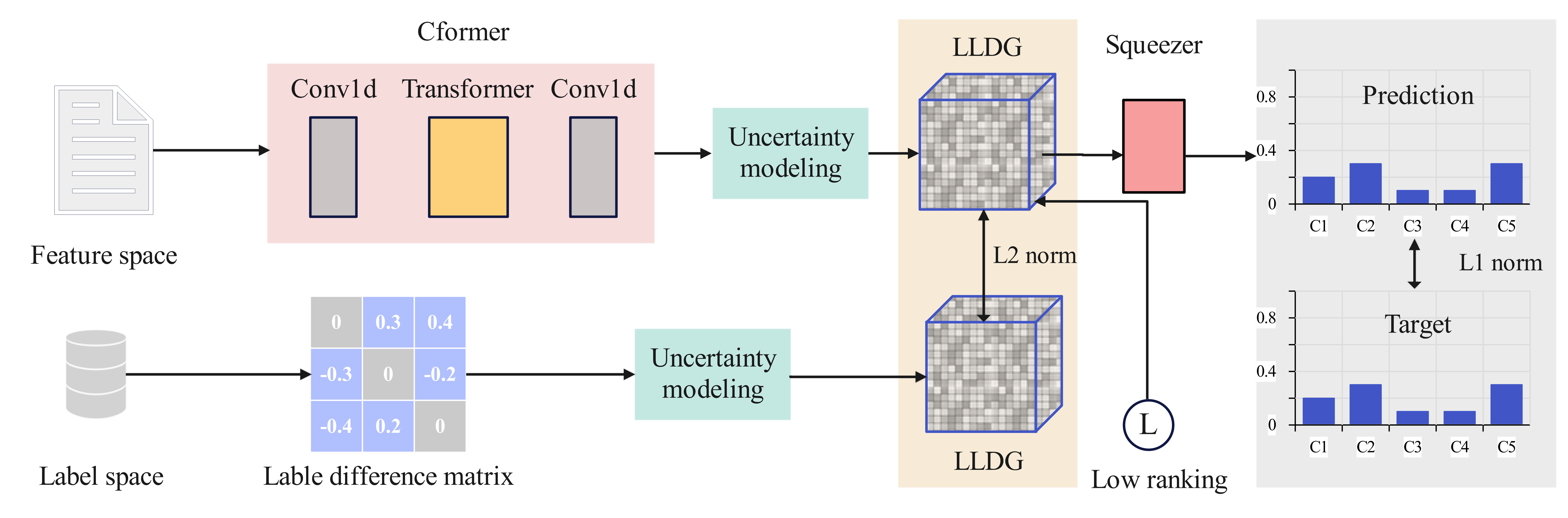}}
	\caption{\textbf{Our architecture}. In general, our model has two learning targets, on the one hand, to learn the label distribution, and on the other hand to learn the LLDG. Specifically, first, the input features are extracted by a local-global feature extractor to create a grid. LLDG establishes a labeled correlation space that is constrained by the Tucker reconstruction algorithm. Then, this grid is conducted in LLDG-Mixer to form a vector by squeezing. Finally, this vector is normalized by \texttt{Softmax} to form a label distribution.}
	\label{fig.1}
	\vspace{-2mm}
\end{figure*}

\section{Method}
As shown in Figure \ref{fig.1} of the framework, this section explains the details of LLDG and the generated LDL.
Latent label distribution grid representation involves notation definition, pipeline description, and loss function.

\noindent \textbf{Notation.}
Let $\mathbf{X}^{\text{m} \times \text{n}}$ denote the feature matrix of the input (customized LDL datasets are all tabular data), where m denotes the number of samples and n denotes the feature space dimension. 
Let $\mathbf{L}^{\text{c}}$ denote the c-dimensional label. 
In LDL, to better describe the correspondence between samples and labels, $\mathbf{L}$ is denoted as $\mathbf{L}_{i} = \{d^{l_j}_{x_i},...,d^{l_c}_{x_i}\}$, where $d^{l_j}_{x_i}$ (abbreviated as $d^j_i$ ) indicates the degree of the description of the j-th label for the i-th sample.
For $d$, there are two constraints $d^j_i \in [0,1]$ and $\sum_{j} d^j_i = 1$  to constrain $\mathbf{L}_{i}$ to be a distribution.
In addition, the LLDG is denoted as $\mathbf{B}$, the difference matrix is denoted as $\mathbf{D}$.
The goal of LDL is to learn the feature of the sample and calculate the degree to which the label describes the sample.

\noindent \textbf{Local-global feature extractor.}
To build a high-quality representation space, we deliberately design a pipeline to generate latent representations (LLDG).
%
%
%
Specifically, to build a qualified LLDG, we propose a CFormer as a feature extractor. 
%
%
First, each sample $x_{i}$ can be deemed as a 1D signal or a sequence of tokens.
Then, we use several 1D convolutional kernels ($\text{Conv}_{1 \times 1}$) to extract the features of the $x_{i}$.
Next, the long-range dependencies of $x_{i}$ are modeled using a standard Transformer~\cite{dosovitskiy2020image}.
Finally, the feature map computed by Transformer is \textbf{d}eformed and \textbf{s}queezed (DS) to yield an $c \times c \times c$ grid $\mathbf{B}$.
The whole workflow can be formulated as follows:
\begin{equation}
    \mathbf{B} = \texttt{Tanh}(\text{DS}(\text{Transformer}(\text{Conv}_{1 \times 1}(\mathbf{X})))), \mathbf{B} \in (-1, 1).
\end{equation}
Note that the convolution kernels in these several 1D convolutional layers utilize an equivariant array \{3, 5, 7\} to capture the local features of the tabular data.
Local-global feature extractor's activation function uses \texttt{PReLU} (\texttt{PReLU} $\in$ (-$\infty$, +$\infty$)) to enforce the nonlinear ability in this network.
In addition, $\mathbf{B}$ is bounded by the \texttt{Tanh} function to avoid outliers due to stochasticity.
Next, we describe the characteristics and optimization schemes of this grid $\mathbf{B}$.

\noindent \textbf{Latent label distribution grid.}
LLDG $\mathbf{B}$ is a latent representation with high capacity and elemental correlation.
To efficiently leverage LLDG $\mathbf{B}$, we conduct two regularization terms on it.
One of them is the label relation and the other is the low-rank characteristics.
Specifically, we first construct a $c \times c$ difference matrix (symmetric matrix) $\mathbf{D}$ in the label space, i.e., the labels compute the difference $a_{ij}$ between each other.
\begin{equation*}
    \mathbf{D} = 
	\begin{bmatrix}
		a_{11} & a_{12} & \cdots & a_{1c} \\
		a_{21} & a_{22} & \cdots & a_{2c} \\
		\vdots & \vdots & \ddots & \vdots \\
		a_{c1} & a_{c2} & \cdots & a_{cc}
	\end{bmatrix}, ~~a_{ij} = d^{j}-d^{i}.
\end{equation*}
Here, we also try the strategy of dividing between label values, but the grid is not built smoothly (the overall learning space is a multi-peaked saddle).
Then, we create a grid $\hat{\mathbf{B}}$ from $\mathbf{D}$ (2D $\longrightarrow$ 3D).
We use $a_{ij}$ as the mean and $1 - \text{abs}(a_{ij})$ (the clearer the relationship between label values, the smaller the uncertainty) as the variance for Gaussian sampling $\mathcal N$ to obtain a distribution of length $c$.
%
%
Each element of the $\mathbf{D}$ matrix conducts this procedure to generate an $c \times c \times c$ grid $\hat{\mathbf{B}}$.
Finally, this $\hat{\mathbf{B}}$ serves as an a priori knowledge to bound $\mathbf{B}$ during training.
The whole workflow can be written:
\begin{equation}
   \underset{\text{g}}{\text{min}} ~\mathbf{B} \xleftarrow[\text{g}]{} \hat{\mathbf{B}},~~~ \hat{\mathbf{B}} \leftarrow \mathcal{N}(\mathbf{D}), 
\end{equation}
g denotes the distance of minimizing two components, and here we use the L1 norm.
Since Gaussian functions are inherently extremely uncertain, an LLDG regularization algorithm with low-rank characteristics is introduced to alleviate the noise caused by sampling.
Unlike algorithms such as SVD and PCA, since $\mathbf{B}$ is a tensor, here we develop the \textbf{Tucker reconstruction algorithm} to enforce a low-rank characteristic on $\mathbf{B}$.
Specifically, the Tucker reconstruction algorithm is a two-stage approach, i.e., Tucker decomposition and recovery.
First,  given any third-order tensor ${\mathcal{Y}}\in\mathbb{R}^{M\times N\times T}$ , if the rank of the Tucker decomposition is assumed to be ($R_1,R_2,R_3$), then for any element $y_{ijt}$, the expression of the Tucker decomposition is
\begin{equation}
\displaystyle{y_{ijt}\approx\sum_{r_1=1}^{R_1}\sum_{r_2=1}^{R_2}\sum_{r_3=1}^{R_3}g_{r_1r_2r_3}u_{ir_1}v_{jr_2}x_{tr_3}},\forall (i,j,t),
\end{equation}
where the core tensor $\mathcal{G}$ has size $R_1 \times R_2 \times R_3$ and its elements are denoted as $g_{r_1r_2r_3}, \forall (r_1,r_2,r_3$); the first factor matrix $\boldsymbol{U}$ has size $M \times R_1$ and its elements are denoted as $u_{ir_1}, \forall (i,r_1)$; the size of the second factor matrix $\boldsymbol{U}$ is $M \times R_1$, and its elements are written as $u_{ir_1},\forall 
(i,r_1)$; the size of the second factor matrix $\boldsymbol{V}$ is $N\times R_2$, and its elements are written as $v_{jr_2}, \forall (j,r_2)$; the size o fthe third factor matrix $\boldsymbol{X}$ is $T\times R_3$ , and its elements are written as $x_{tr_3}$, $\forall (t,r_3)$.
The expression for Tucker's decomposition can also be written
\begin{equation}
\boldsymbol{\mathcal{Y}}\approx\boldsymbol{\mathcal{G}}\times_1\boldsymbol{U}\times_2\boldsymbol{V}\times_3\boldsymbol{X},
\end{equation}
where the symbol $\times_k,k=1,2,3$ represents the product between tensors and matrices, which is also called the modal product.
After that, the Tucker recovery process can be viewed as the inverse of the above process with the help of the obtained core tensor and factor matrices.
This regularization algorithm with low-rank characteristics can be written as:
\begin{equation}
    \mathbf{B}^{*} = \text{Tucker[R]}(\text{Tucker[D]}(\mathbf{B})),
\end{equation}
where Tucker[D] and Tucker[R] represent the processes of decomposition and recovery, respectively.
$\mathbf{B}^{*}$ denotes the tensor that has the low-rank characteristics enforced by the Tucker reconstruction.

\noindent \textbf{Generate label distribution.}
Since there is an indirect correlation between the labels, we attempt to conduct global modeling on $\mathbf{B}^{*}$ to regress an accurate label distribution.
For this purpose, we develop a \textbf{LLDG-Mixer} that recreates a label distribution from $\mathbf{B}$.
%
During the reconstruction process, LLDG-Mixer first uses three linear layers (the number of neurons is $c$) to learn the x-axis, y-axis, and z-axis information in the grid $\mathbf{B}$, respectively.
Note that the dimensions of the grid $\mathbf{B}$ are swapped between these three linear layers.
Next, the grid $\mathbf{B}$ is squeezed into the z-axis using an adding operator ($ao^z$) to yield a map.
Then, this map is squeezed on the x-axis via the adding operator ($ao^x$) to generate a vector.
Finally, this vector is normalized by the \texttt{Softmax} function to generate the label distribution.
LLDG-Mixer models long-range dependencies in a ``rolling'' fashion across the three dimensions of the grid, formalized as follows:
\begin{equation}
    \mathbf{L} = \texttt{softmax}(ao^x(ao^z(\text{MLP}(\mathbf{B})))),
\end{equation}
where MLP represents three linear layers, LLDG-Mixer uses \texttt{ReLU} as the activation function.

\noindent \textbf{Loss function.}
The loss function (Total\_loss) consists of two parts: grid loss $\text{Loss}_{g}$ and label distribution loss $\text{Loss}_{d}$.
%
%
We denote the predicted label distribution as $\hat{L}$ and the real label distribution as $L$.
%
\begin{equation}
    \text{Loss}_{d} = \frac{1}{c}\sum_{i=1}^{c}||\hat{L}_i-L_i||^2,
\end{equation}
\vspace{-1mm}
\begin{equation}
    \text{Loss}_{g} = \frac{1}{c^3}\sum_{i,j,k=1}^{c}||B_{ijk}-\hat{B}_{ijk})||^2,
\end{equation}
\vspace{-1mm}
\begin{equation}
    \text{Total\_loss} = \text{Loss}_{d}+ \lambda \times \text{Loss}_{g},
\end{equation}
where $\lambda$ denotes the weight, and here we use 0.5.
In addition, we also try KL divergence and other regularization schemes, and they do not work significantly.
Referring to the design of the variational autoencoder, we try to conduct a standard normal distribution on $\mathbf{B}$ to remove the instability caused by the variance, but its effectiveness is not satisfactory.

\section{Experiment}
The purpose of this section is to evaluate the effectiveness of the LLDG. The experiments are organized into two main parts, one for evaluation using a customized LDL dataset, and the other for using LLDG as a representation space to conduct the classification task on a publicly available benchmark.
In addition, we visualize the change of the LLDG with the enhancement of the label space noise, thus demonstrating the robustness of our method.

\noindent \textbf{Datasets.}
We conduct experiments on 15 LDL datasets to evaluate the model's performance in terms of accuracy. 
These datasets include the biological, image (human face), and text (movie) domains. 
The SJAFEE dataset uses 243-dimensional features (a pre-trained ResNet18/50 is employed to extract features) to represent the facial characteristics of a Japanese female. 
For each sample, 60 experts score six emotions (happiness, sadness, surprise, fear, anger, and disgust), and the normalized average score is used as the label distribution corresponding to the sample.
The Movie dataset contains 7755 movie ratings on Netflix. 
The Movie dataset uses a 1,869-dimensional feature vector (a pre-trained BERT is employed to extract features) to represent a movie and a 5-dimensional label distribution as the learning target, where the label distribution comes from the audience's evaluation of the movie.
%
%
The Human Gene dataset has the most examples on the customized LDL dataset, each of which contains 36 features and the corresponding 68 diseases.
wc-LDL is a dataset containing many watercolors with characteristics reported in~\cite{zheng2022label}; SBU-3DFE is a face dataset made in the same process as the construction of SJAFFE.
The remaining 12 datasets can be found at \footnote{\url{http://palm.seu.edu.cn/xgeng/LDL/index.htm}}.
There are datasets summarized in Table~\ref{tab:datasets}.

\begin{minipage}{\textwidth} \scriptsize 
		\centering 
	\begin{minipage}[h]{0.480\textwidth}
		\flushleft
		\makeatletter\def\@captype{table}\makeatother
        \begin{tabular}{llccc}
        \toprule
        Index  & Data set &  Instances &  Features & Labels \\
        \midrule
        1 & Yeast-alpha & 2465 & 24 & 18        \\
        2 & Yeast-cdc & 2465 & 24 & 15         \\
        3 & Yeast-cold & 2465 & 24 & 4         \\
        4 & Yeast-diau & 2465 & 24 & 14         \\
        5 & Yeast-dtt & 2465 & 24 & 4         \\
        6 & Yeast-elu & 2465 & 24 & 14         \\
        7 & Yeast-heat & 2465 & 24 & 6         \\
        8 & Yeast-spo & 2465 & 24 & 6         \\
        9 & Yeast-spoem & 2465 & 24 & 2         \\
        10 & Yeast-spo5 & 2465 & 24 & 3           \\
        11 & SJAFFE & 213 & 243 & 6               \\
        12 & Human Gene & 30542 & 36 & 68         \\
        13 & Movie & 7755 & 1869 & 5         \\
        14 & wc-LDL  &  500        & 243       & 12  \\
        15 & SBU-3DFE  &  2500      & 243       & 6   \\
        \bottomrule
	    \end{tabular}
	    \caption{The 15 datasets include detailed statistics of instances, features, and labels.}
	    \vspace{0mm}
	    \label{tab:datasets}
	\end{minipage}
	\begin{minipage}[h]{0.480\textwidth}
	\flushright
	\makeatletter\def\@captype{table}\makeatother

	    \begin{tabular}{c|c|c}
		\toprule
		
		Name & Formula &Value field\\
		\midrule
		$\text{Chebyshev} \downarrow$ & $\max_j{\lvert d_j-\hat{d_j} \rvert}$ & (0, 1)  \\
		\midrule
		$\text{Clark} \downarrow$ & $\sqrt{\sum_{j=1}^{L} \frac{(d_j-\hat{d_j})^2}{(d_j+\hat{d_j})^2}}$  & (0, +$\infty$)\\
		\midrule
		$\text{K-L} \downarrow$ & $\sum_{j=1}^{L} d_j \ln\frac{d_j}{\hat{d_j}}$  & (0, +$\infty$)\\
		\midrule
		$\text{Canberra} \downarrow$ & $\sum_{j=1}^{L} \frac{\lvert d_j-\hat{d_j} \rvert}{d_j+\hat{d_j}}$  & (0, +$\infty$) \\
		\midrule
		$\text{Cosine} \uparrow$& $\frac{\sum_{j=1}^{L}d_j^2 \hat{d_j}^2}{\sqrt{\sum_{j=1}^{L}d_j^2} \sqrt{\sum_{j=1}^{L}\hat{d}_j^2}}$ & (0, 1)\\
		\midrule
		$\text{Intersection} \uparrow$&$\sum_{j=1}^{L}\min(d_j,\hat{d_j})$ & (0, 1) \\
		\bottomrule
	\end{tabular}
	\caption{Evaluation metrics for LDL algorithms, where $\uparrow$ and $\downarrow$ represents performance favorites.}
	\vspace{0mm}
	\label{tab:metrics}
	\end{minipage}
	\end{minipage}

\noindent \textbf{Evaluation measures.}
To objectively evaluate the performance of the algorithm, we use six metrics proposed by Geng~\cite{geng2016label}, which are Chebyshev distance (Chebyshev $\downarrow$), KL distances (K-L $\downarrow$), Canberra distance (Canberra $\downarrow$), Clark distance (Clark $\downarrow$), Intersection similarity (Intersection $\uparrow$) and Cosine similarity (Cosine $\uparrow$).
Their calculation formulas are as shown in Table~\ref{tab:metrics}, where $\downarrow$ represents the smaller the better, and $\uparrow$ represents the bigger the better. 
In addition, we estimate the value domain of each metric conducting a pair of label distributions\footnote[1]{Predicting label distribution and ground truth, each label distribution strictly follows the definition of a label distribution (the sum of the label distribution values is 1 and each label distribution value belongs to [0, 1]).} to build an effective \textit{moat}.
%

%


\noindent \textbf{Experiment settings.}
Seven LDL algorithms are compared with our method, including AA-KNN, AA-BP, IIS-LLD, BFGS-LLD, Duo-LDL, ENC-BFGS, and ENC-SDPT3. 
AA-BP and AA-KNN are two transformation algorithms. The three-layer MLP and KNN algorithms are modified to adapt to the LDL task.
IIS-LLD and BFGS-LLD are methods with the same loss function but different optimization algorithms. 
In addition, the same is true for ENC-BFGS and ENC-SDPT3. 
Duo-LDL is an improvement of the AA-BP method. 
Although it is also a three-layer MLP structure, the output dimension of the last layer has changed from c to c $\times$ (c-1), and the learning target has also changed from label to label difference.

%
Except for wc-LDL, the evaluation results of all algorithms are extracted from the existing paper (all data sets are divided into training and test sets in a ratio of 8:2).
For our approach, we use the PyTorch 1.7 platform, an RTX3090 GPU; the AdamW optimizer is deployed on all datasets to learn the weights of the model with a learning rate of 0.001.
To evaluate the soundness of the design of the optimization objective, we use wandb\footnote{\url{https://github.com/wandb/wandb}} to estimate the importance of the loss function term.
The results show that the average contribution of $\text{Loss}_{d}$ and $\text{Loss}_{g}$ is 82.3\% and 17.7\% respectively across the 15 data sets.

\begin{table*}[!htb]\tiny
	\centering
	\caption{Experimental results on 9 datasets and the best results are bolded.}
	\resizebox{\textwidth}{!}{
		\begin{tabular}{c|c|cccccc}
			\toprule
			Dataset                             & Algorithm	   & Chebyshev $\downarrow$  & Clark $\downarrow$   &Canberra $\downarrow$   &K-L $\downarrow$    &Cosine $\uparrow$   &Intersection $\uparrow$           \\
			\midrule
			\multirow{8}[2]{*}{Yeast-alpha}
			
			&AA-KNN	    &0.0147 $\pm$ 0.0008			&0.2320 $\pm$ 0.0120		&0.7582 $\pm$ 0.0289	&0.0064 $\pm$ 0.0004	&0.9938 $\pm$ 0.0003	&0.9581 $\pm$ 0.0021 \\
			&AA-BP	    &0.0401 $\pm$ 0.0022  		&0.7110 $\pm$ 0.0540		&2.3521 $\pm$ 0.1282	&0.0771 $\pm$ 0.0084	&0.9391 $\pm$ 0.0059	&0.8772 $\pm$ 0.0081 \\
			&IIS-LLD	&0.0156 $\pm$ 0.0004			&0.2330 $\pm$ 0.0120		&0.7625 $\pm$ 0.0351	&0.0075 $\pm$ 0.0003	&0.9927 $\pm$ 0.0003	&0.9578 $\pm$ 0.0021 \\
			&Duo-LDL	&0.0160 $\pm$ 0.0010			&0.2100 $\pm$ 0.0120		&0.6813 $\pm$ 0.0186	&0.0056 $\pm$ 0.0004	&0.9946 $\pm$ 0.0003	&0.9624 $\pm$ 0.0009 \\
			&BFGS-LLD	&0.0135 $\pm$ 0.0003			&0.2100 $\pm$ 0.0140		&0.6845 $\pm$ 0.0170	&0.0063 $\pm$ 0.0003	&0.9943 $\pm$ 0.0003	&0.9621 $\pm$ 0.0009 \\
			&ENC-BFGS	&\textbf{0.0134 $\pm$ 0.0004}	&	-				&0.6812 $\pm$ 0.0174	&0.0056 $\pm$ 0.0004	&0.9946 $\pm$ 0.0003	&0.9624 $\pm$ 0.0009  \\
			&ENC-SDPT3	&\textbf{0.0134 $\pm$ 0.0004}	& -					&0.6813 $\pm$ 0.0171	&0.0056 $\pm$ 0.0004 	&0.9946 $\pm$ 0.0003	&0.9624 $\pm$ 0.0009 \\
			&Ours		&\textbf{0.0134 $\pm$ 0.0004}	&\textbf{0.2082 $\pm$ 0.0054}	&\textbf{0.6767 $\pm$ 0.1680}	&\textbf{0.0054 $\pm$ 0.0003}	&\textbf{0.9947 $\pm$ 0.0003}	&\textbf{0.9626 $\pm$ 0.0009} \\

			\midrule
			\multirow{8}[2]{*}{Yeast-cdc}
			&AA-KNN		&0.0173 $\pm$ 0.0004	&0.2370 $\pm$ 0.0140	&0.7172 $\pm$ 0.0215	&0.0083 $\pm$ 0.0005	&0.9924 $\pm$ 0.0004	&0.9528 $\pm$ 0.0028 \\
			&AA-BP		&0.0410 $\pm$ 0.0022  &0.5680 $\pm$ 0.0370	&1.7152 $\pm$ 0.1055	&0.0608 $\pm$ 0.0065	&0.9508 $\pm$ 0.0047	&0.8912 $\pm$ 0.0051 \\
			&IIS-LLD	&0.0184 $\pm$ 0.0005	&0.2350 $\pm$ 0.0120	&0.7086 $\pm$ 0.0215	&0.0092 $\pm$ 0.0005	&0.9915 $\pm$ 0.0004	&0.9532 $\pm$ 0.0021 \\
			&Duo-LDL	&0.0165 $\pm$ 0.0007	&0.2160 $\pm$ 0.0013	&0.6462 $\pm$ 0.0180	&0.0074 $\pm$ 0.0005	&0.9933 $\pm$ 0.0003	&0.9575 $\pm$ 0.0010 \\
			&BFGS-LLD	&0.0164 $\pm$ 0.0008	&0.2160 $\pm$ 0.0130	&0.6487 $\pm$ 0.0161	&0.0074 $\pm$ 0.0005	&0.9928 $\pm$ 0.0004	&0.9574 $\pm$ 0.0010 \\
			&ENC-BFGS	&0.0162 $\pm$ 0.0005	&-				&0.6461 $\pm$ 0.0173	&0.0074 $\pm$ 0.0005	&0.9933 $\pm$ 0.0003	&0.9575 $\pm$ 0.0010 \\
			&ENC-SDPT3	&0.0162 $\pm$ 0.0005	&-				&0.6466 $\pm$ 0.0157 	&0.0074 $\pm$ 0.0005	&0.9933 $\pm$ 0.0003	&0.9575 $\pm$ 0.0010 \\
			&Ours		&\textbf{0.1587 $\pm$ 0.0004}	&\textbf{0.2130 $\pm$ 0.0044}	&\textbf{0.6445 $\pm$ 0.0089}	&\textbf{0.0068 $\pm$ 0.0002}	&\textbf{0.9934 $\pm$ 0.0003}	&\textbf{0.9576 $\pm$ 0.0006} \\
			
			\midrule
			\multirow{8}[2]{*}{Yeast-cold}
			&AA-KNN	&0.0542 $\pm$ 0.0017	&0.1500 $\pm$ 0.0070	&0.2604 $\pm$ 0.0102	&0.0142 $\pm$ 0.0018	&0.9872 $\pm$ 0.0008	&0.9362 $\pm$ 0.0024 \\
			&AA-BP		&0.0598 $\pm$ 0.0031  &0.1550 $\pm$ 0.0090	&0.2681 $\pm$ 0.0143	&0.0174 $\pm$ 0.0026	&0.9844 $\pm$ 0.0016	&0.9340 $\pm$ 0.0032 \\
			&IIS-LLD	&0.0545 $\pm$ 0.0017	&0.1440 $\pm$ 0.0050	&0.2487 $\pm$ 0.0091	&0.0144 $\pm$ 0.0020	&0.9871 $\pm$ 0.0009	&0.9376 $\pm$ 0.0022 \\
			&Duo-LDL	&0.0512 $\pm$ 0.0015	&0.1410 $\pm$ 0.0050	&0.2408 $\pm$ 0.0092	&0.0129 $\pm$ 0.0020	&\textbf{0.9886 $\pm$ 0.0008} 	&0.9409 $\pm$ 0.0019 \\
			&BFGS-LLD	&0.0513 $\pm$ 0.0016	&0.1390 $\pm$ 0.0050	&0.2402 $\pm$ 0.0090	&0.0130 $\pm$ 0.0014	&0.9880 $\pm$ 0.0007	&0.9408 $\pm$ 0.0019 \\
			&ENC-BFGS	&0.0510 $\pm$ 0.0018	&-				&0.2398 $\pm$ 0.0089	&0.0129 $\pm$ 0.0022	&\textbf{0.9886 $\pm$ 0.0008}	&0.9409 $\pm$ 0.0021 \\
			&ENC-SDPT3	&0.0510 $\pm$ 0.0017	&-				&0.2398 $\pm$ 0.0080	&0.0129 $\pm$ 0.0018	&\textbf{0.9886 $\pm$ 0.0007}	&0.9409 $\pm$ 0.0019 \\
			&Ours		&\textbf{0.0508 $\pm$ 0.0021}	&\textbf{0.1385 $\pm$ 0.0054}	&\textbf{0.2386 $\pm$ 0.0090}	&\textbf{0.0121 $\pm$ 0.0010}	&\textbf{0.9886 $\pm$ 0.0008}	&\textbf{0.9411 $\pm$ 0.0021} \\
			
			\midrule
			\multirow{8}[2]{*}{Yeast-diau}
			&AA-KNN		&0.0385 $\pm$ 0.0012	&0.2120 $\pm$ 0.0040	&0.4551 $\pm$ 0.0112	&0.0151 $\pm$ 0.0012	&0.9866 $\pm$ 0.0008	&0.9371 $\pm$ 0.0021 \\
			&AA-BP		&0.0531 $\pm$ 0.0053  &0.2630 $\pm$ 0.0170	&0.5675 $\pm$ 0.0310	&0.0299 $\pm$ 0.0069	&0.9742 $\pm$ 0.0053	&0.9224 $\pm$ 0.0039 \\
			&IIS-LLD	0&.0397 $\pm$ 0.0011	&0.2090 $\pm$ 0.0070	&0.4487 $\pm$ 0.0170	&0.0159 $\pm$ 0.0011	&0.9861 $\pm$ 0.0007	&0.9381 $\pm$ 0.0020 \\
			&Duo-LDL	&0.0370 $\pm$ 0.0012	&0.2020 $\pm$ 0.0060	&0.4331 $\pm$ 0.0100	&0.0138 $\pm$ 0.0010	&0.9879 $\pm$ 0.0007	&0.9402 $\pm$ 0.0017 \\
			&BFGS-LLD	&0.0374 $\pm$ 0.0011	&\textbf{0.2000 $\pm$ 0.0090}	&0.4312 $\pm$ 0.0103	&0.0139 $\pm$ 0.0012	&0.9869 $\pm$ 0.0007	&\textbf{0.9403 $\pm$ 0.0017} \\
			&ENC-BFGS	&0.0370 $\pm$ 0.0012	&-				&0.4312 $\pm$ 0.0140	&0.0140 $\pm$ 0.0011	&0.9879 $\pm$ 0.0007	&0.9402 $\pm$ 0.0017 \\ 
			&ENC-SDPT3	&0.0370 $\pm$ 0.0012	&-				&0.4311 $\pm$ 0.0129	&0.0140 $\pm$ 0.0011	&0.9879 $\pm$ 0.0007	&\textbf{0.9403 $\pm$ 0.0017} \\
			&Ours		&\textbf{0.0368 $\pm$ 0.0010}	&\textbf{0.2000 $\pm$ 0.0045}	&\textbf{0.4307 $\pm$ 0.0109}	&\textbf{0.0130 $\pm$ 0.0006}	&\textbf{0.9880 $\pm$ 0.0005}	&0.9402 $\pm$ 0.0015 \\
			
			\midrule
			\multirow{8}[2]{*}{Yeast-dtt}
			&AA-KNN		&0.0385 $\pm$ 0.0013	&0.1060 $\pm$ 0.0040	&0.1821 $\pm$ 0.0071	&0.0076 $\pm$ 0.0016	&0.9933 $\pm$ 0.0005	&0.9549 $\pm$ 0.0021 \\
			&AA-BP		&0.0470 $\pm$ 0.0042  &0.1180 $\pm$ 0.0070	&0.2043 $\pm$ 0.0118	&0.0114 $\pm$ 0.0027	&0.9898 $\pm$ 0.0021	&0.9502 $\pm$ 0.0021 \\
			&IIS-LLD	&0.0406 $\pm$ 0.0014	&0.1050 $\pm$ 0.0040	&0.1812 $\pm$ 0.0051	&0.0084 $\pm$ 0.0016	&0.9926 $\pm$ 0.0005	&0.9552 $\pm$ 0.0016 \\
			&Duo-LDL	&0.0361 $\pm$ 0.0012	&0.0980 $\pm$ 0.0039	&0.1689 $\pm$ 0.0060 	&0.0068 $\pm$ 0.0013	&0.9941 $\pm$ 0.0004	&0.9582 $\pm$ 0.0014 \\
			&BFGS-LLD	&0.0360 $\pm$ 0.0013	&0.0980 $\pm$ 0.0040	&0.1690 $\pm$ 0.0062	&0.0071 $\pm$ 0.0015	&0.9940 $\pm$ 0.0005	&0.9583 $\pm$ 0.0015 \\
			&ENC-BFGS	&0.0359 $\pm$ 0.0012	&-				&0.1687 $\pm$ 0.0065	&0.0069 $\pm$ 0.0015	&0.9941 $\pm$ 0.0004	&0.9584 $\pm$ 0.0014 \\
			&ENC-SDPT3	&0.0359 $\pm$ 0.0012	&-				&0.1686 $\pm$ 0.0062	&0.0069 $\pm$ 0.0015	&0.9941 $\pm$ 0.0005	&0.9584 $\pm$ 0.0014 \\
			&Ours		&\textbf{0.0356 $\pm$ 0.0014}	&\textbf{0.0967 $\pm$ 0.0043}	&\textbf{0.1680 $\pm$ 0.0077}	&\textbf{0.0059 $\pm$ 0.0006}	&\textbf{0.9942 $\pm$ 0.0005}	&\textbf{0.9586 $\pm$ 0.0019} \\
			
			\midrule
			\multirow{8}[2]{*}{Yeast-elu}
			&AA-KNN		&0.0173 $\pm$ 0.0004	&0.2180 $\pm$ 0.0050	&0.6442 $\pm$ 0.0143	&0.0073 $\pm$ 0.0004	&0.9931 $\pm$ 0.0002	&0.9546 $\pm$ 0.0011 \\
			&AA-BP		&0.0409 $\pm$ 0.0023  &0.5050 $\pm$ 0.0340	&1.4885 $\pm$ 0.0672	&0.0540 $\pm$ 0.0058	&0.9557 $\pm$ 0.0042	&0.8992 $\pm$ 0.0054 \\
			&IIS-LLD	&0.0186 $\pm$ 0.0004	&0.2160 $\pm$ 0.0070	&0.6387 $\pm$ 0.0193	&0.0083 $\pm$ 0.0004	&0.9922 $\pm$ 0.0003	&0.9547 $\pm$ 0.0011 \\
			&Duo-LDL	&0.0164 $\pm$ 0.0005	&0.2110 $\pm$ 0.0059	&0.5853 $\pm$ 0.0115	&0.0065 $\pm$ 0.0004	&0.9940 $\pm$ 0.0002	&0.9591 $\pm$ 0.0010 \\
			&BFGS-LLD	&0.0165 $\pm$ 0.0004	&0.1990 $\pm$ 0.0050	&0.5831 $\pm$ 0.0142	&0.0066 $\pm$ 0.0005	&0.9940 $\pm$ 0.0003	&0.9589 $\pm$ 0.0009 \\
			&ENC-BFGS	&0.0163 $\pm$ 0.0004	&-				&0.5823 $\pm$ 0.0128	&0.0064 $\pm$ 0.0004	&0.9941 $\pm$ 0.0002	&0.9589 $\pm$ 0.0015 \\
			&ENC-SDPT3	&0.0163 $\pm$ 0.0004	&-				&0.5825 $\pm$ 0.0130	&0.0064 $\pm$ 0.0004 	&0.9940 $\pm$ 0.0003	&0.9589 $\pm$ 0.0009 \\
			&Ours		&\textbf{0.0160 $\pm$ 0.0004}	&\textbf{0.1965 $\pm$ 0.0036}	&\textbf{0.5801 $\pm$ 0.0070}	&\textbf{0.0061 $\pm$ 0.0002}	&\textbf{0.9942 $\pm$ 0.0002}	&\textbf{0.9592 $\pm$ 0.0007} \\
			\midrule
			\multirow{8}[2]{*}{Yeast-heat}
			&AA-KNN		&0.0441 $\pm$ 0.0012		&0.1950 $\pm$ 0.0050	&0.3918 $\pm$ 0.0112	&0.0145 $\pm$ 0.0011	&0.9867 $\pm$ 0.0006	&0.9362 $\pm$ 0.0023 \\
			&AA-BP		&0.0534 $\pm$ 0.0035   	&0.2280 $\pm$ 0.0150	&0.4589 $\pm$ 0.0286	&0.0244 $\pm$ 0.0038	&0.9782 $\pm$ 0.0030	&0.9251 $\pm$ 0.0054 \\
			&IIS-LLD	&0.0495 $\pm$ 0.0013		&0.1880 $\pm$ 0.0030	&0.3772 $\pm$ 0.0068	&0.0156 $\pm$ 0.0012	&0.9857 $\pm$ 0.0007	&0.9384 $\pm$ 0.0011 \\
			&Duo-LDL	&0.0422 $\pm$ 0.0013		&0.1831 $\pm$ 0.0031	&0.3646 $\pm$ 0.0100	&0.0133 $\pm$ 0.0013	&\textbf{0.9880 $\pm$ 0.0006}	&\textbf{0.9406 $\pm$ 0.0014} \\
			&BFGS-LLD	&0.0425 $\pm$ 0.0013		&0.1820 $\pm$ 0.0030	&0.3642 $\pm$ 0.0072	&0.0135 $\pm$ 0.0012	&0.9878 $\pm$ 0.0006	&0.9402 $\pm$ 0.0016 \\
			&ENC-BFGS	&0.0422 $\pm$ 0.0012		&-				&0.3642 $\pm$ 0.0090	&0.0133 $\pm$ 0.0011	&\textbf{0.9880 $\pm$ 0.0006}	&0.9402 $\pm$ 0.0014 \\
			&ENC-SDPT3	&0.0422 $\pm$ 0.0012		&-				&0.3640 $\pm$ 0.0098	&0.0133 $\pm$ 0.0010	&\textbf{0.9880 $\pm$ 0.0006}	&0.9403 $\pm$ 0.0016 \\
			&Ours		&\textbf{0.0421 $\pm$ 0.0010}		&\textbf{0.1819 $\pm$ 0.0040}	&\textbf{0.3631 $\pm$ 0.0076}	&\textbf{0.0129 $\pm$ 0.0009}	&\textbf{0.9880 $\pm$ 0.0008}	&\textbf{0.9406 $\pm$ 0.0020} \\
			
			\midrule
			\multirow{8}[2]{*}{Yeast-spo}
			&AA-KNN		&0.0627 $\pm$ 0.0023	&0.2710 $\pm$ 0.0110	&0.5597 $\pm$ 0.0218	&0.0303 $\pm$ 0.0021	&0.9730 $\pm$ 0.0017	&0.9082 $\pm$ 0.0043  \\
			&AA-BP		&0.0684 $\pm$ 0.0031  &0.2920 $\pm$ 0.0220	&0.5992 $\pm$ 0.0417	&0.0368 $\pm$ 0.0037	&0.9679 $\pm$ 0.0029	&0.9022 $\pm$ 0.0069  \\
			&IIS-LLD	&0.0605 $\pm$ 0.0018	&0.2550 $\pm$ 0.0170	&0.5231 $\pm$ 0.0312	&0.0281 $\pm$ 0.0019	&0.9753 $\pm$ 0.0013	&0.9143 $\pm$ 0.0052 \\
			&Duo-LDL	&0.0585 $\pm$ 0.0020	&0.2530 $\pm$ 0.0168	&0.5137 $\pm$ 0.0135	&0.0258 $\pm$ 0.0017	&0.9770 $\pm$ 0.0012	&0.9156 $\pm$ 0.0023 \\
			&BFGS-LLD	&0.0586 $\pm$ 0.0021	&0.2500 $\pm$ 0.0170	&0.5133 $\pm$ 0.0145	&0.0265 $\pm$ 0.0018	&0.9768 $\pm$ 0.0013	&0.9155 $\pm$ 0.0023 \\
			&ENC-BFGS	&0.0583 $\pm$ 0.0018	&-				&0.5127 $\pm$ 0.0155	&0.0263 $\pm$ 0.0017	&0.9770 $\pm$ 0.0012	&0.9156 $\pm$ 0.0023 \\
			&ENC-SDPT3	&0.0583 $\pm$ 0.0018	&-				&0.5126 $\pm$ 0.0144	&0.0263 $\pm$ 0.0018 	&0.9770 $\pm$ 0.0012	&0.9156 $\pm$ 0.0023 \\
			&Ours		&\textbf{0.0582 $\pm$ 0.0012}	&\textbf{0.2489 $\pm$ 0.0062}	&\textbf{0.5124 $\pm$ 0.0164}	&\textbf{0.0246 $\pm$ 0.0013}	&\textbf{0.9771 $\pm$ 0.0012}	&\textbf{0.9160 $\pm$ 0.0023} \\
			
			\midrule
			\multirow{8}[2]{*}{Yeast-spoem}
			&AA-KNN		&0.0904 $\pm$ 0.0047	&0.1370 $\pm$ 0.0060	&0.1914 $\pm$ 0.0089	&0.0291 $\pm$ 0.0037	&0.9764 $\pm$ 0.0023	&0.9072 $\pm$ 0.0043 \\
			&AA-BP		&0.0892 $\pm$ 0.0049  &0.1323 $\pm$ 0.0080	&0.1842 $\pm$ 0.0108	&0.0283 $\pm$ 0.0034	&0.9778 $\pm$ 0.0034	&0.9108 $\pm$ 0.0056 \\
			&IIS-LLD	&0.0905 $\pm$ 0.0036	&0.1321 $\pm$ 0.0070	&0.1840 $\pm$ 0.0099	&0.0291 $\pm$ 0.0035	&0.9774 $\pm$ 0.0015	&0.9109 $\pm$ 0.0054 \\
			&Duo-LDL	&0.0871 $\pm$ 0.0037	&0.1290 $\pm$ 0.0080	&0.1812 $\pm$ 0.0072	&0.0255 $\pm$ 0.0030	&0.9790 $\pm$ 0.0015	&0.9128 $\pm$ 0.0058 \\
			&BFGS-LLD	&0.0870 $\pm$ 0.0037	&0.1290 $\pm$ 0.0080	&0.1799 $\pm$ 0.0082	&0.0270 $\pm$ 0.0035	&0.9790 $\pm$ 0.0015	&0.9131 $\pm$ 0.0038 \\
			&ENC-BFGS	&0.0873 $\pm$ 0.0037	&-				&0.1808 $\pm$ 0.0092	&0.0273 $\pm$ 0.0037	&0.9788 $\pm$ 0.0016	&0.9127 $\pm$ 0.0042 \\
			&ENC-SDPT3	&0.0874 $\pm$ 0.0037	&-				&0.1808 $\pm$ 0.0082	&0.0273 $\pm$ 0.0038	&0.9788 $\pm$ 0.0016	&0.9126 $\pm$ 0.0037 \\
			&Ours		&\textbf{0.0865 $\pm$ 0.0055}	&\textbf{0.1278 $\pm$ 0.0090}	&\textbf{0.1780 $\pm$ 0.0096}	&\textbf{0.0250 $\pm$ 0.0027}	&\textbf{0.9791 $\pm$ 0.0017}	&\textbf{0.9150 $\pm$ 0.0038} \\
			\bottomrule
		\end{tabular}%
	}
	\label{tab:result1}%
\end{table*}%
\begin{table*}[!htb]\tiny
	\centering
	\caption{Experimental results on 6 datasets and the best results are bolded.}
	\resizebox{\textwidth}{!}{
		\begin{tabular}{c|c|cccccc}
			\toprule
			Dataset                             & Algorithm	   & Chebyshev $\downarrow$  & Clark $\downarrow$   &Canberra $\downarrow$   &K-L $\downarrow$    &Cosine $\uparrow$   &Intersection $\uparrow$           \\
			\midrule
			\multirow{8}[2]{*}{Yeast-spo5}
			&AA-KNN		&0.0948 $\pm$ 0.0036	&0.1930 $\pm$ 0.0110	&0.2969 $\pm$ 0.0146	&0.0343 $\pm$ 0.0031	&0.9713 $\pm$ 0.0022	&0.9044 $\pm$ 0.0051 \\
			&AA-BP		&0.0949 $\pm$ 0.0036  &0.1890 $\pm$ 0.0120	&0.2912 $\pm$ 0.0170	&0.0339 $\pm$ 0.0032	&0.9723 $\pm$ 0.0019	&0.9062 $\pm$ 0.0054 \\
			&IIS-LLD	&0.0931 $\pm$ 0.0037	&0.1870 $\pm$ 0.0130	&0.2871 $\pm$ 0.0191	&0.0330 $\pm$ 0.0032	&0.9731 $\pm$ 0.0019	&0.9072 $\pm$ 0.0034 \\
			&Duo-LDL	&0.0913 $\pm$ 0.0033	&0.1840 $\pm$ 0.0124	&0.2821 $\pm$ 0.0100	&0.0293 $\pm$ 0.0022	&0.9741 $\pm$ 0.0018	&0.9088 $\pm$ 0.0033 \\
			&BFGS-LLD	&0.0914 $\pm$ 0.0040	&0.1840 $\pm$ 0.0120	&0.2829 $\pm$ 0.0101	&0.0324 $\pm$ 0.0031	&0.9741 $\pm$ 0.0016	&0.9086 $\pm$ 0.0031 \\
			&ENC-BFGS	&0.0912 $\pm$ 0.0038	&-				&0.2823 $\pm$ 0.0115	&0.0322 $\pm$ 0.0034	&0.9741 $\pm$ 0.0018	&0.9088 $\pm$ 0.0034 \\
			&ENC-SDPT3	&0.0912 $\pm$ 0.0038	&-				&0.2824 $\pm$ 0.0104	&0.0322 $\pm$ 0.0034	&0.9741 $\pm$ 0.0016	&0.9088 $\pm$ 0.0033 \\
			&Ours		&\textbf{0.0906 $\pm$ 0.0028}	&\textbf{0.1828 $\pm$ 0.0058}	&\textbf{0.2814 $\pm$ 0.0070}	&\textbf{0.0286 $\pm$ 0.0024}	&\textbf{0.9742 $\pm$ 0.0019}	&\textbf{0.9089 $\pm$ 0.0036} \\

			\midrule
			\multirow{8}[2]{*}{SJAFFE}
			&AA-KNN		&0.1141 $\pm$ 0.0108	&0.4100 $\pm$ 0.0500	&0.8431 $\pm$ 0.1131	&0.0712 $\pm$ 0.0231	&0.9337 $\pm$ 0.0182 	&0.8552 $\pm$ 0.0215 \\
			&AA-BP		&0.1272 $\pm$ 0.0126  &0.5100 $\pm$ 0.0540	&1.0462 $\pm$ 0.1250	&0.0960 $\pm$ 0.0183	&0.9145 $\pm$ 0.0140	&0.8243 $\pm$ 0.0216 \\
			&IIS-LLD	&0.1194 $\pm$ 0.0130	&0.4190 $\pm$ 0.0340	&0.8751 $\pm$ 0.0842	&0.0700 $\pm$ 0.0089	&0.9314 $\pm$ 0.0104	&0.8513 $\pm$ 0.0147 \\
			&Duo-LDL	&0.1291 $\pm$ 0.0120	&0.5080 $\pm$ 0.0350	&0.8142 $\pm$ 0.0700	&0.1061 $\pm$ 0.0112	&0.9100 $\pm$ 0.0100	&0.8310 $\pm$ 0.0123 \\ 
			&BFGS-LLD	&0.1142 $\pm$ 0.0132	&0.3990 $\pm$ 0.0440	&0.8202 $\pm$ 0.0675	&0.0740 $\pm$ 0.0135	&0.9301 $\pm$ 0.0121	&0.8606 $\pm$ 0.0121 \\
			&ENC-BFGS	&0.0956 $\pm$ 0.0103	&-				&\textbf{0.7108 $\pm$ 0.0553}	&0.0500 $\pm$ 0.0090	&\textbf{0.9531 $\pm$ 0.0086} 	&\textbf{0.8797 $\pm$ 0.0101} \\
			&ENC-SDPT3	&0.0959 $\pm$ 0.0103	&-				&0.7115 $\pm$ 0.0612	&0.0500 $\pm$ 0.0090	&0.9530 $\pm$ 0.0090	&\textbf{0.8797 $\pm$ 0.0108} \\
			&Ours		&\textbf{0.0933 $\pm$ 0.0067}	&\textbf{0.3545 $\pm$ 0.0018}	&0.7193 $\pm$ 0.0455	&\textbf{0.0494 $\pm$ 0.0055}	&0.9529 $\pm$ 0.0053	&0.8789 $\pm$ 0.0082 \\

			\midrule
			\multirow{8}[2]{*}{Human gene}
			&AA-KNN		&0.0648 $\pm$ 0.0019	&2.3880 $\pm$ 0.1090	&16.2832 $\pm$ 0.8072	&0.3010 $\pm$ 0.0084	&0.7687 $\pm$ 0.0046	&0.7433 $\pm$ 0.0128 \\
			&AA-BP		&0.0624 $\pm$ 0.0019  &3.3440 $\pm$ 0.2500	&22.7847 $\pm$ 1.8523	&0.4691 $\pm$ 0.0169	&0.6906 $\pm$ 0.0087	&0.6712 $\pm$ 0.0221 \\
			&IIS-LLD	&0.0534 $\pm$ 0.0016	&2.1230 $\pm$ 0.0880	&14.5412 $\pm$ 0.6534	&0.2440 $\pm$ 0.0035	&0.8334 $\pm$ 0.0040	&0.7828 $\pm$ 0.0098 \\
			&Duo-LDL	&0.0534 $\pm$ 0.0007	&2.1100 $\pm$ 0.0880	&14.4423 $\pm$ 0.2176	&0.2358 $\pm$ 0.0110	&0.8345 $\pm$ 0.0020	&0.7852 $\pm$ 0.0042 \\
			&BFGS-LLD	&0.0534 $\pm$ 0.0018	&2.1110 $\pm$ 0.0860	&14.4532 $\pm$ 0.2207	&0.2480 $\pm$ 0.0015	&0.8342 $\pm$ 0.0039	&0.7846 $\pm$ 0.0034 \\
			&ENC-BFGS	&0.0534 $\pm$ 0.0018	&-				&14.4543 $\pm$ 0.2282	&0.2264 $\pm$ 0.0072	&0.8342 $\pm$ 0.0039	&0.7846 $\pm$ 0.0028 \\
			&ENC-SDPT3	&0.0533 $\pm$ 0.0018	&-				&14.4543 $\pm$ 0.2282	&\textbf{0.2262 $\pm$ 0.0072}	&0.8345 $\pm$ 0.0039	&0.7842 $\pm$ 0.0034 \\
			&Ours		&\textbf{0.0522 $\pm$ 0.0011}	&\textbf{2.1008 $\pm$ 0.0252}	&\textbf{14.3775 $\pm$ 0.1721}	&0.2313 $\pm$ 0.0054	&\textbf{0.8368 $\pm$ 0.0027}	&\textbf{0.7856 $\pm$ 0.0024} \\
			
			\midrule
			\multirow{8}[2]{*}{Movie}
			&AA-KNN		&0.1542 $\pm$ 0.0048	&0.6520 $\pm$ 0.0230	&1.2758 $\pm$ 0.0457	&0.2008 $\pm$ 0.0102	&0.8802 $\pm$ 0.0026	&0.7801 $\pm$ 0.0056 \\
			&AA-BP		&0.1572 $\pm$ 0.0024  &0.6750 $\pm$ 0.0480	&1.2693 $\pm$ 0.0872	&0.1792 $\pm$ 0.0246	&0.8948 $\pm$ 0.0012	&0.7882 $\pm$ 0.0112 \\
			&IIS-LLD	&0.1508 $\pm$ 0.0016	&0.5910 $\pm$ 0.0280	&1.1367 $\pm$ 0.0542	&0.1368 $\pm$ 0.0121	&0.9067 $\pm$ 0.0023	&0.8004 $\pm$ 0.0100 \\
			&Duo-LDL	&0.1240 $\pm$ 0.0032	&0.5750 $\pm$ 0.0290	&1.0772 $\pm$ 0.0201	&0.1131 $\pm$ 0.0625	&0.9264 $\pm$ 0.0032	&0.8221 $\pm$ 0.0040 \\
			&BFGS-LLD	&0.1355 $\pm$ 0.0018	&0.5890 $\pm$ 0.0380	&1.0617 $\pm$ 0.0173	&0.1292 $\pm$ 0.0056	&0.9231 $\pm$ 0.0028	&0.8192 $\pm$ 0.0054 \\
			&ENC-BFGS	&0.1199 $\pm$ 0.0256	&-				&1.0345 $\pm$ 0.0195	&0.1210 $\pm$ 0.0049	&0.9298 $\pm$ 0.0027	&0.8282 $\pm$ 0.0034 \\
			&ENC-SDPT3	&0.1197 $\pm$ 0.0024	&-				&\textbf{1.0337 $\pm$ 0.0175}	&0.1209 $\pm$ 0.0049	&\textbf{0.9299 $\pm$ 0.0032}	&\textbf{0.8284 $\pm$ 0.0032} \\
			&Ours		&\textbf{0.1196 $\pm$ 0.0016}	&\textbf{0.5735 $\pm$ 0.0045}	&1.0339 $\pm$ 0.0280	&\textbf{0.1074 $\pm$ 0.0057}	&0.9290 $\pm$ 0.0035	&0.8258 $\pm$ 0.0042 \\
			\midrule
			\multirow{5}[2]{*}{wc-LDL}
			&  AA-KNN    & 0.1122 $\pm$ \text{0.0039}           & 1.5657 $\pm$ \text{0.0021}       & 0.7998 $\pm$ \text{0.0020}          & 0.0498 $\pm$ \text{0.0051}     & 0.9704 $\pm$ \text{0.0036}      & 0.8611 $\pm$ \text{0.0016}
			\\ 
			&  AA-BP       & 0.0989 $\pm$ \text{0.0019}          & 0.6689 $\pm$ \text{0.0019}       & 0.8089 $\pm$ \text{0.0049}          & 0.0477 $\pm$ \text{0.0018}     & 0.9476 $\pm$ \text{0.0020}      & 0.8700 $\pm$ \text{0.0033}                    \\
			&  IIS-LLD       & 0.1009 $\pm$ \text{0.0038}           & 0.4199 $\pm$ \text{0.0044}       & 0.9008 $\pm$ \text{0.0015}          & 0.0519 $\pm$ \text{0.0040}     & 0.9779 $\pm$ \text{0.0018}      & 0.8660 $\pm$ \text{0.0022}                     \\
			&  Duo-LDL    & 0.1057 $\pm$ \text{0.0019}           & 1.0569 $\pm$ \text{0.0039}       & 0.7890 $\pm$ \text{0.0039}          & 0.0545 $\pm$ \text{0.0037}     & 0.9668 $\pm$ \text{0.0049}      & 0.8383 $\pm$ \text{0.0018}                     \\
			&  BFGS-LLD     & 0.0923 $\pm$ \text{0.0030}          & 0.4212 $\pm$ \text{0.0036}      & 0.8135 $\pm$ \text{0.0024}          & 0.0511 $\pm$ \text{0.0049}     & 0.9718 $\pm$ \text{0.0022}      & 0.8669 $\pm$ \text{0.0047}                   \\
			&     Ours     & \textbf{0.0745 $\pm$ 0.0012}           & \textbf{0.3684 $\pm$ 0.0055}       & \textbf{0.7660 $\pm$ 0.0033}          & \textbf{0.0419 $\pm$ 0.0008}     & \textbf{0.9897 $\pm$ 0.0009}      & \textbf{0.8819 $\pm$ 0.0014}                \\
			\midrule
			\multirow{5}[2]{*}{SBU-3DFE}
			&  AA-KNN    & 0.1119 $\pm$ \text{0.0030}           & 1.4657 $\pm$ \text{0.0022}       & 0.7700 $\pm$ \text{0.0025}          & 0.0492 $\pm$ \text{0.0053}     & 0.9753 $\pm$ \text{0.0036}      & 0.8710 $\pm$ \text{0.0019}
			\\
			&  AA-BP       & 0.0899 $\pm$ \text{0.0021}           & 0.6563 $\pm$ \text{0.0019}       & 0.8132 $\pm$ \text{0.0100}          & 0.0468 $\pm$ \text{0.0021}     & 0.9441 $\pm$ \text{0.0011}      & 0.8723 $\pm$ \text{0.0034}                    \\
			&  IIS-LLD       & 0.1009 $\pm$ \text{0.0038}           & 0.4199 $\pm$ \text{0.0044}       & 0.9008 $\pm$ \text{0.0015}          & 0.0519 $\pm$ \text{0.0040}     & 0.9780 $\pm$ \text{0.0029}      & 0.8660 $\pm$ \text{0.0022}                     \\
			&  Duo-LDL       & 0.1009 $\pm$ \text{0.0038}           & 0.4199 $\pm$ \text{0.0044}       & 0.9008 $\pm$ \text{0.0015}          & 0.0519 $\pm$ \text{0.0040}     & 0.9780 $\pm$ \text{0.0029}      & 0.8660 $\pm$ \text{0.0022}                     \\
			&  BFGS-LLD    & 0.1100 $\pm$ \text{0.0025}           & 0.9660 $\pm$ \text{0.0039}       & 0.7897 $\pm$ \text{0.0033}          & 0.0511 $\pm$ \text{0.0021}     & 0.9677 $\pm$ \text{0.0056}      & 0.8555 $\pm$ \text{0.0032}                   \\
			&  Ours        & \textbf{0.0811 $\pm$ 0.0023}           & \textbf{0.3987 $\pm$ 0.0024}       & \textbf{0.7533 $\pm$ 0.0027}          & \textbf{0.0354 $\pm$ 0.0031}      & \textbf{0.9888 $\pm$ 0.0066}      & \textbf{0.8997 $\pm$ 0.0030}                    \\
			\bottomrule
		\end{tabular}%
	}
	\label{tab:result2}%
\end{table*}%

\noindent \textbf{Results.}
As shown in Table~\ref{tab:result1} and~\ref{tab:result2}of experimental results, the best results are marked in bold.
Since ENC-SDPT3 and ENC-BFGS do not publish their codes, there are no Clark metrics for the two methods. 
%
%
It is worth comparing LLDG, AA-BP, and Duo-LDL because all three are neural network-based algorithms. 
As can be seen from the table, LLDG outperforms both on most datasets. 
Furthermore, it can be seen from Table~\ref{tab:result2} that ENC-SDPT3 and ENC-BFGS achieve comparable performance to our method on Movie and SJAFFE. 
This may be due to interfering features in these two datasets with large feature dimensions, making these two feature reconstruction-based methods also able to play their roles.

\noindent \textbf{Ablation study.}
To verify the role of LLDG, we conduct ablation experiments on the Gene dataset. 
Specifically, we modify the output dimension of LLDG-Mixer and remove the $\text{Loss}_{g}$ from the loss function.
In addition, to demonstrate the effectiveness of the Tucker reconstruction algorithm, we just remove the Tucker reconstruction algorithm using a raw LLDG.
The experimental results are summarized in Table~\ref{tab:ablation}, showing the effectiveness of Grid.
\begin{table}[htbp] \scriptsize 
	\vspace{0mm}
	\centering
    \begin{tabular}{c|c|c|c}
        \toprule

			Measures & w/ LLDG & w/o LLDG & w/o Tucker\\
			\midrule
			$\text{Chebyshev} \downarrow$ & \textbf{0.0522 $\pm$ 0.0011} & 0.0524 $\pm$ 0.0009 & 0.0529 $\pm$ 0.0011\\
			$\text{Clark} \downarrow$ & \textbf{2.1008 $\pm$ 0.0252} & 2.1189 $\pm$  0.0177 & 2.1180 $\pm$  0.0172\\
			$\text{K-L} \downarrow$ & \textbf{0.2313 $\pm$ 0.0054} & 0.2333 $\pm$ 0.0035  & 0.2334 $\pm$ 0.0039\\
			$\text{Canberra} \downarrow$ & \textbf{14.3775 $\pm$ 0.1721} & 14.4975 $\pm$  0.1331 & 14.4905 $\pm$  0.1202\\
			$\text{Cosine} \uparrow$ & \textbf{0.8368 $\pm$ 0.0027} & 0.8346 $\pm$ 0.0019  & 0.8349 $\pm$ 0.0008\\
			$\text{Intersection} \uparrow$ & \textbf{0.7856 $\pm$ 0.0024} & 0.7840 $\pm$ 0.0017
   & 0.7842 $\pm$ 0.0019\\
        \bottomrule
    \end{tabular}

\label{tab:ablation}%
\caption{Results on ablation studies. This table demonstrates the role of LLDG, and it is worth noting that the Tucker reconstruction algorithm has significant benefits in boosting the performance of the algorithm. More ablation results are shown in the Appendix.}
\vspace{-0mm}
\end{table}

\noindent \textbf{Noise disturbance.}
To verify the robustness of LLDG, we test the model with different degrees of Gaussian noise. 
The standard deviation of the noise is \{0.1, 0.2, 0.5, 1.0\}, the mean is 0, and the experimental results are shown in Table~\ref{tab:noise}.
Although the model's performance continues to decline as the standard deviation increases, it is still competitive compared to other methods.
LLDG shows that the model is robust to noise.
\begin{table}[htbp] \scriptsize 
	\vspace{0mm}
	\centering
	
		\begin{tabular}{c|c|c|c|c}
			\toprule
			Measures
			 & \multicolumn{1}{c|}{0.1} & \multicolumn{1}{c|}{0.2} & \multicolumn{1}{c|}{0.5} & \multicolumn{1}{c}{1.0}\\
		\midrule
		$\text{Chebyshev} \downarrow$ & 0.0524 $\pm$ 0.0017 & 0.0528 $\pm$ 0.0019 & 0.0526 $\pm$ 0.0012 & 0.0531 $\pm$ 0.0015\\
		
		$\text{Clark} \downarrow$ & 2.1102 $\pm$ 0.0194 & 2.1113 $\pm$ 0.0120 & 2.1214 $\pm$ 0.0202 & 2.1243 $\pm$ 0.0202\\
		$\text{K-L} \downarrow$ & 0.2330 $\pm$ 0.0070 & 0.2335 $\pm$ 0.00755 & 0.2359 $\pm$ 0.0064 & 0.2362 $\pm$ 0.0060\\
		
		$\text{Canberra} \downarrow$ & 14.4344 $\pm$ 0.01591 & 14.4346 $\pm$ 0.01967 & 14.5237 $\pm$ 0.01783 & 14.5489 $\pm$ 0.1715\\
		$\text{Cosine} \uparrow$& 0.8360 $\pm$ 0.0037 & 0.8353 $\pm$ 0.0013 & 0.8355 $\pm$ 0.0030 & 0.8338 $\pm$ 0.0036\\
		$\text{Intersection} \uparrow$& 0.7852 $\pm$ 0.0025 & 0.7849 $\pm$ 0.0020 & 0.7841 $\pm$ 0.0028 & 0.7833 $\pm$ 0.0027\\
		\bottomrule
	\end{tabular}%

\label{tab:noise}%
\caption{Results on noise interference. Statistically, LLDG shows an outstanding performance even when a Gaussian noise with a variance of 1 is added to the label distribution space.}
\vspace{-0mm}
\end{table}
\noindent \textbf{Potential of LLDG.}
Our model can be extended to handle classification.
For every single label, by modeling the Gaussian prior is expanded into a vector shape as a learning target for the LLDG.
We conduct experiment reports to show the potential of the public dataset.
We evaluate the algorithm's performance on the MedMNIST Classification Decathlon benchmark.
The area under the ROC curve (AUC) and Accuracy (ACC) is used as the evaluation metrics.
Our model is trained for 100 epochs, using a cross-entropy loss and an AdamW optimizer with a batch size of 128 and an initial learning rate of $1 \times 10 ^{-3}$.
The overall performance of the methods is reported in Table~\ref{tab:Results}.
Table~\ref{tab:Results} shows that our method achieves competitive results.
\begin{table*}[!htb] \scriptsize
	\caption{{\bf Overall performance of MedMNIST} in metrics of AUC and ACC, using ResNet-18 / ResNet-50 with resolution $28$ and $224$, auto-sklearn , AutoKeras and Google AutoML Vision.}
	\label{tab:Results}
	\vspace{-1mm}
	\begin{center}
		
		\begin{tabular}{@{}ccccccccccc@{}}
			\toprule
			\multirow{2}{*}{Methods} &
			\multicolumn{2}{c}{PathMNIST} &
			\multicolumn{2}{c}{ChestMNIST} &
			\multicolumn{2}{c}{DermaMNIST} &
			\multicolumn{2}{c}{OCTMNIST} &
			\multicolumn{2}{c}{PneumoniaMNIST} \\
			& AUC & ACC & AUC & ACC & AUC & ACC & AUC & ACC & AUC & ACC \\ \midrule
			ResNet-18 (28)     & 0.972 & 0.844 & 0.706 & 0.947 & 0.899 & 0.721 & 0.951 & \bf0.758 & 0.957 & 0.843  \\
			ResNet-18 (224)        & 0.978 & 0.860 & 0.713 & \bf0.948 & 0.896 & 0.727 & 0.960 & 0.752 & 0.970 & 0.861  \\
			ResNet-50 (28)        & 0.979 & \bf0.864 & 0.692 & 0.947 & 0.886 & 0.710 & 0.939 & 0.745 & 0.949 & 0.857  \\
			ResNet-50 (224)       & 0.978 & 0.848 & 0.706 & 0.947 & 0.895 & 0.719 & 0.951 & 0.750 & 0.968 & 0.896  \\
			auto-sklearn         & 0.500 & 0.186 & 0.647 & 0.642 & 0.906 & 0.734 & 0.883 & 0.595 & 0.947 & 0.865  \\
			AutoKeras           & 0.979 & \bf0.864 & 0.715 & 0.939 & 0.921 & 0.756 & 0.956 & 0.736 & 0.970 & 0.918 \\
			Google AutoML Vision  & 0.982 & 0.811 & 0.718 & 0.947 & 0.925 & \bf0.766 & 0.965 & 0.732 & \bf0.993 & 0.941 \\
			Ours  & \bf0.986& \bf0.877 & \bf0.723 & \bf0.949 & \bf0.933 & 0.765 & \bf0.969 & 0.756 & 0.993 & \bf0.949 \\
			\midrule
			\multirow{2}{*}{Methods} &
			\multicolumn{2}{c}{RetinaMNIST} &
			\multicolumn{2}{c}{BreastMNIST} &
			\multicolumn{2}{c}{OrganMNIST\_A} &
			\multicolumn{2}{c}{OrganMNIST\_C} &
			\multicolumn{2}{c}{OrganMNIST\_S} \\
			& AUC & ACC & AUC & ACC & AUC & ACC & AUC & ACC & AUC & ACC \\ \midrule
			ResNet-18 (28)         & 0.727 & 0.515 & 0.897 & 0.859 & 0.995 & 0.921 & 0.990 & 0.889 & 0.967 & 0.762 \\
			ResNet-18 (224)       & 0.721 & 0.543 & 0.915 & 0.878 & \bf0.997 & 0.931 & 0.991 & 0.907 & \bf0.974 & 0.777 \\
			ResNet-50 (28)         & 0.719 & 0.490 & 0.879 & 0.853 & 0.995 & 0.916 & 0.990 & 0.893 & 0.968 & 0.746 \\
			ResNet-50 (224)        & 0.717 & 0.555 & 0.863 & 0.833 & \bf0.997 & 0.931 & \bf0.992 & 0.898 & 0.970 & 0.770 \\
			auto-sklearn        & 0.694 & 0.525 & 0.848 & 0.808 & 0.797 & 0.563 & 0.898 & 0.676 & 0.855 & 0.601 \\
			AutoKeras           & 0.655 & 0.420 & 0.833 & 0.801 & 0.996 & 0.929 & \bf0.992 & 0.915 & 0.972 & 0.803 \\
			Google AutoML Vision  & \bf0.762 & 0.530    &  \bf0.932 & 0.865    & 0.988 & 0.818 & 0.986 & 0.861 & 0.964 & 0.706 \\
			Ours                  & 0.761    & \bf0.572 & 0.931     & \bf0.893 & 0.996 & \bf0.934 & 0.990 & \bf0.919 & 0.972 & \bf0.823 \\
			\bottomrule
		\end{tabular}
		\vspace{-2mm}
	\end{center}
\end{table*}

%
%
%
\begin{figure*}[t]
	\centerline{\includegraphics[width=0.80\textwidth]{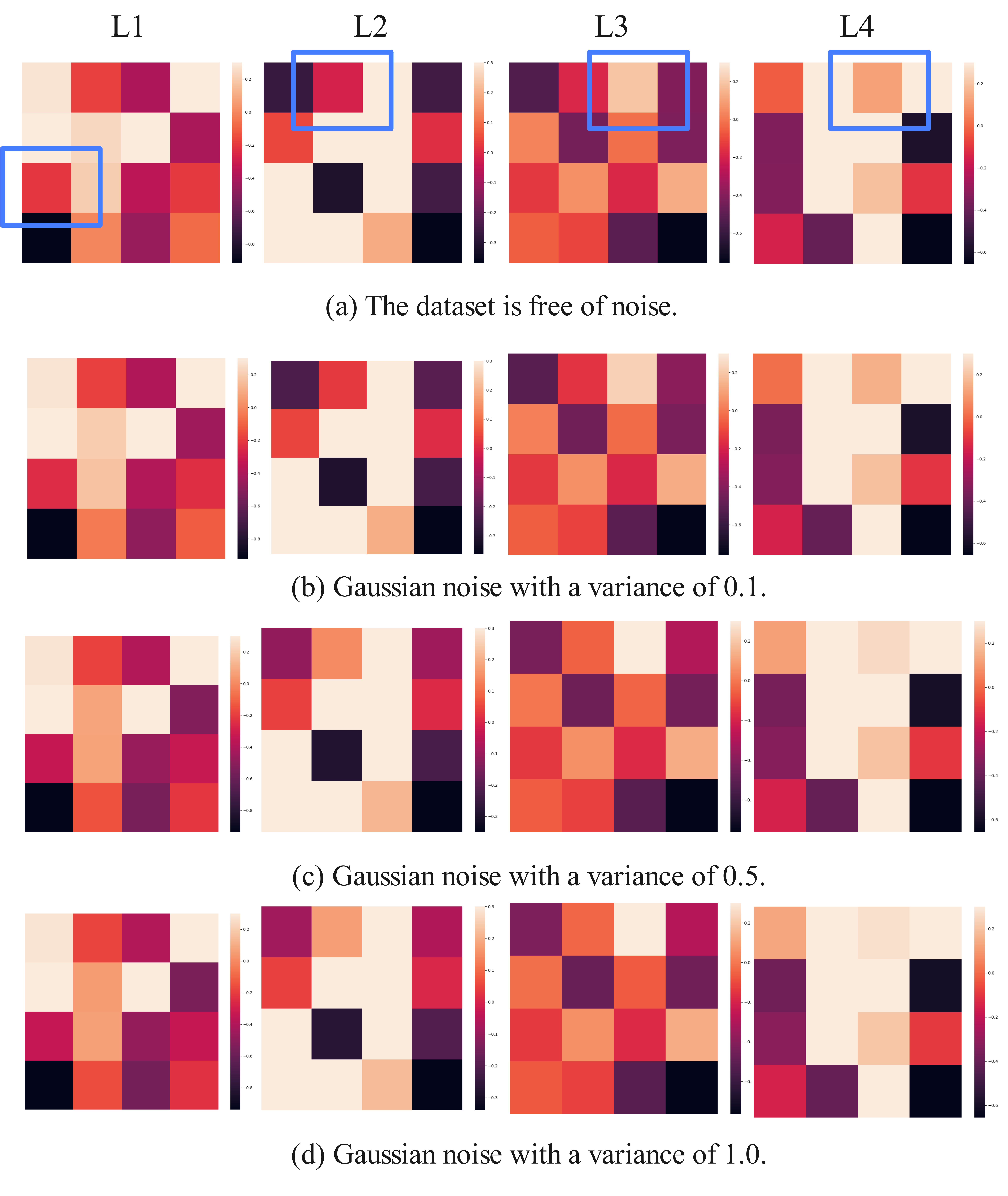}}
	\caption{This figure shows the energy distribution of the LLDG (4), and the blue boxes indicate the regions where the energy varies with the enhancement of Gaussian noise; in general, the energy of the LLDG does not vary significantly with the increase of Gaussian noise. }
	\label{fig-Vis}
	\vspace{-0mm}
\end{figure*}
\noindent \textbf{Visualization and analysis.}
To demonstrate the anti-noise ability of the model more intuitively, we visualize the impact of different noise levels on the LLDG (our method is conducted on Yest-dtt, where the size of the grid is a $4^{3}$), as shown in the Figure~\ref{fig-Vis}.
It is known from the observation that LLDG still has a nearly consistent representation space under different levels of noise interference.
In addition, we statistically measure the distribution of the parameters of the network generating LLDG, and the results show that the parameters of the network show a standard normal distribution.
In contrast, the parameter distribution of the model without LLDG is a Gaussian distribution with a mean of 0.2.
Overall, the strong robustness of our method comes from two aspects: 1) the parameter space of the model is relatively stable, and 2) the Tucker reconstruction technique eliminates the noise.

\section{Limitations and Broad Impact}
Although LLDG has good potential for uncertainty modeling, there are two limitations of our approach.
a) A large number of computing resources are consumed: when the dimensionality of the label distribution space is large (Human Gene dataset), our method takes 4 hours to train an epoch on a single RTX3090 GPU.
b) For an arbitrary dataset of classification tasks, logical labels or integer-type labels require empirical construction into a label distribution, which often needs a lot of work for trial and error. 
In this paper, the datasets involved are publicly released by the researchers, and the LLDG algorithm does not touch on the issue of personal privacy and security.

\section{Conclusion}
In this paper, we introduce a latent representation with uncertainty (LLDG) to address the problem of uncertainty in the label space of an arbitrary dataset.
The effectiveness of LLDG modeling comes from the uncertainty based on Gaussian sampling and the regularization characteristics of the Tucker reconstruction technique.
Numerous experiments validate that LLDG shows amazing potential in noise-inclusive, noise-free LDL datasets.
In addition, our method is extended to a classification task still showing competitive results.
\clearpage
\bibliography{egbib}

\clearpage
\section*{Checklist}

Please do not modify the questions and only use the provided macros for your
answers.  Note that the Checklist section does not count toward the page
limit.  In your paper, please delete this instructions block and only keep the
Checklist section heading above along with the questions/answers below.

\end{document}